# Beyond Multiple-Choice Accuracy: Real-World Challenges of Implementing Large Language Models in Healthcare

Yifan Yang[1], Qiao Jin[1], Qingqing Zhu[1], Zhizheng Wang[1], Francisco Erramuspe Álvarez[1], Nicholas Wan[1], Benjamin Hou[1], and Zhiyong Lu[1,*]

[1]National Library of Medicine (NLM), National Institutes of Health (NIH), Bethesda, MD 20894, USA

[*]Correspondence: zhiyong.lu@nih.gov

## Abstract

Large Language Models (LLMs) have gained significant attention in the medical domain for their human-level capabilities, leading to increased efforts to explore their potential in various healthcare applications. However, despite such a promising future, there are multiple challenges and obstacles that remain for their real-world uses in practical settings. This work discusses key challenges for LLMs in medical applications from four unique aspects: operational vulnerabilities, ethical and social considerations, performance and assessment difficulties, and legal and regulatory compliance. Addressing these challenges is crucial for leveraging LLMs to their full potential and ensuring their responsible integration into healthcare.

## Introduction

Large Language Models (LLMs) have emerged as powerful tools in medical applications, offering unprecedented capabilities to process complex medical data, assist in decision-making, and streamline workflows[1–6]. Despite their immense potential, LLMs also present challenges that must be addressed to ensure their safe and effective integration into real-world clinical practice. These challenges range from technical issues such as hallucinations to ethical concerns around data privacy, fairness, and bias. As LLMs continue to being integrated into medical applications, it is essential to understand and address these challenges to utilize LLMs' capabilities effectively while minimizing potential risks.

Unlike LLM applications in other domains, deploying LLMs in medical settings likely requires more caution because patients' lives are at stake. For instance, an erroneous recommendation from an LLM could lead to misdiagnosis or inappropriate treatment, resulting in death of patients[7]. In addition to the technical challenges, deploying LLMs in medicine must also meet more stringent legal and regulatory requirements than general domains because medical applications directly impact patient safety, involve sensitive health data protected by privacy laws, and require compliance with strict standards for clinical accuracy and ethical responsibility to avoid harm or misdiagnosis[8].

Many existing works either focus on summarizing various applications of medical LLMs[2] and/or discuss one or two specific challenges and problems of LLMs in medicine[9–13]. Differently in this work, we aim to aggregate the challenges of applying LLMs from both general and medical domains, using medical-specific examples to provide a more comprehensive and relevant perspective. Figure 1 illustrates the seven specific types of challenges discussed in this work, categorized into four key areas: performance and evaluation challenges, operational vulnerabilities, ethical and social considerations, and legal considerations. Operational vulnerabilities focus on malicious manipulation and hallucination, which pose significant risks to patient safety and the integrity of clinical decision-making. The ethical and social considerations emphasize data privacy and security, alongside fairness and bias, highlighting the need for responsible and equitable use of LLMs in healthcare settings. Performance and assessment challenges include model generalization and evaluation difficulties, which create obstacles to ensuring that LLMs perform reliably across diverse clinical scenarios. Finally, legal considerations pertain to laws and liabilities that are essential for effectively integrating LLMs into clinical practice.

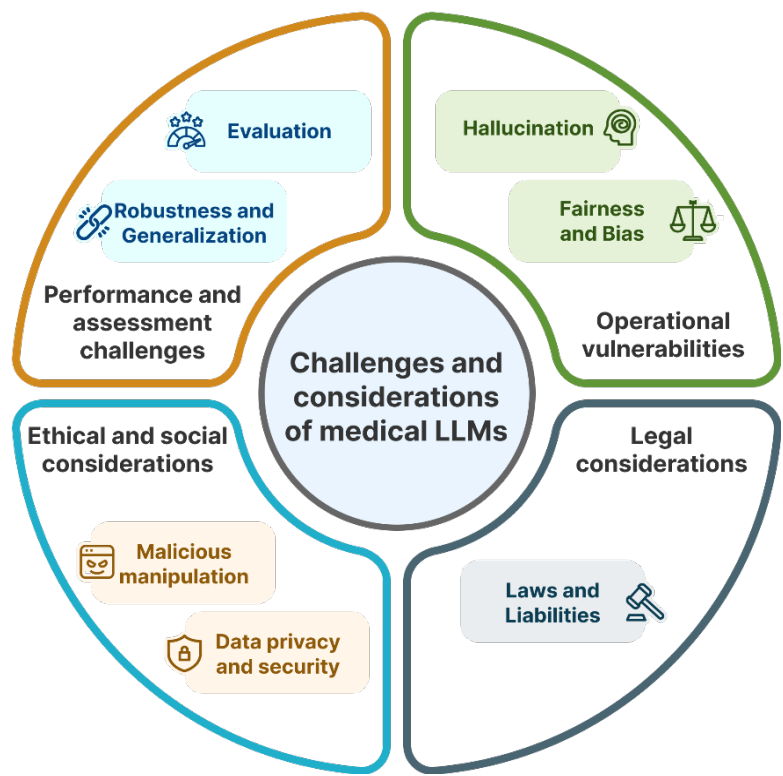

Figure 1. Key challenges and considerations of medical LLMs.

## Performance and evaluation challenges

Currently, most LLMs are evaluated on multi-choice questions (MCQs)[14–17] as a proxy for their medical capabilities, such as the United States Medical Licensing Examination (USMLE) subset of MedQA[18], PubMedQA[19], MedMCQA[20], and medical subsets of MMLU[21]. These datasets are often used because the evaluation can be automatically performed by comparing the predicted answer choice with the ground truth at scale, without requiring any domain expertise. As can be seen in Figure 2, many LLMs have exhibited high performance on such benchmarks. However, there are significant limitations of evaluating LLMs using MCQs, which are unrealistic since no choices will be available in the real-world clinical setting. Evaluation frameworks that mimic physician-patient interactions, such as AgentClinic[22] and Articulate Medical Intelligence Explorer (AMIE)[23], represent a promising future direction to explore.

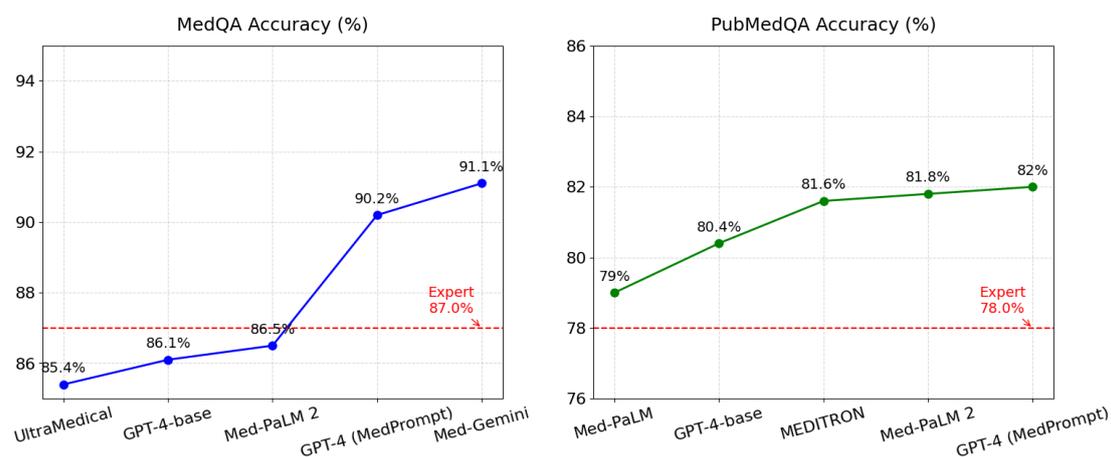

Figure 2. Top-performing models on MedQA (left) and PubMedQA (right) exceed human-expert performance.

Additionally, there can be flaws hidden behind high MCQ scores, where the model predicts correct choices but presents wrong rationales[13]. Specifically, Jin et al evaluated the rationales of GPT-4 Vision for answering medical challenge questions from the NEJM Image Challenge[13]. They found that while GPT-4 Vision achieved expert-level performance measure by multi-choice selection accuracy, the model frequently presents flawed rationales even when it chooses the correct final answer. Such flawed rationales are most common in image comprehension, followed by step-by-step reasoning and the recall of medical knowledge, appearing in roughly 30% of the correct answers.

While MCQ datasets do not reflect real-world tasks, they might still be useful as a screening tool – for example, if a model cannot even pass MedQA-USMLE with a 60% accuracy, the model might not be further considered for any downstream clinical evaluation. This is similar to the screening utility of medical examinations in real life.

Instead of MCQs, real-world evaluations of LLMs on clinical utility often require open-ended questions and answers with real patient information as input[24,25]. In such evaluation scenarios, expert annotations are the ground truth, but obtaining them is time-consuming and labor-intensive, sometimes prohibitively expensive. Moreover, comparing human-annotated answers and LLM-generated text is a non-trivial task. Traditional automated metrics like BLEU[26], ROUGE[27] as well the semantic scores such as BERTScore[28], which focus on word overlap or general meaning, fail to align well with expert judgments because they capture surface-level similarities rather than the critical clinical reasoning and necessary details in medical contexts. As a result, these metrics are not reliable for assessing the accuracy and clinical relevance of AI-generated medical reports[29,30]. As such, there is a need to design novel means to better measure the differences between human-annotated ground truth and LLM-generated text.

Recent work has explored methods that integrate radiologist expertise[31] with LLMs, such as GPT-4[32], using In-Context Instruction Learning (ICIL)[33] and Chain of Thought (CoT)[34] reasoning. These techniques, which involve providing detailed instructions and examples in the input prompt, enable LLMs to evaluate radiology reports in a manner that more closely aligns with the standards of human radiologists. A study showed that GPT-4, guided by these methods, matched expert radiologists' performance while offering a more accurate, context-aware evaluation compared to traditional metrics like BLEU, METEOR, and ROUGE. However, the method relies on expert-generated prompts and annotations, which limits its generalizability. Additionally, it still faces challenges in fully capturing the nuanced clinical context while maintaining scalability and consistency across diverse medical cases[35].

### Model Robustness and Generalization

Model generalizability is a machine learning model's ability to perform well on similar data that is from a different source[36]. In medical AI, this is important because models may be deployed across diverse patient populations, imaging devices, and clinical settings. However, data heterogeneity poses a significant challenge: medical images can vary widely due to differences in equipment and imaging protocols[37], as well as patient cohorts' characteristics like age, ethnicity, and health conditions[38–40].

Typically, models trained on one dataset demonstrate excellent performance on that specific dataset but often fail to generalize to other datasets with different characteristics[41]. General medical LLMs tend to excel in broad domains like MedQA[18] but often underperform on specialized tasks. Meditron achieves 70.2% on MedQA, but achieves lower accuracy in identifying oligometastatic non-small cell lung cancer from radiology text[42]. Meditron's summarization performance is also lower on datasets such as MIMIC-CXR[43], a radiology dataset of x-ray interpretation and reports, and MIMIC-IV[44], an ICU dataset containing ultrasound, CT, and MRI reports[45].

These results highlight the challenge that general medical LLMs face in handling specialized medical subdomains like radiology and ICU care. Hence, fine-tuning these models on domain-specific data is essential to enhance their performance in these specialized areas[46,47]. For example, in ophthalmology-related patient queries, a fine-tuned GPT-3.5 achieved a score of 87.1%, while Llama2-13b scored 80.9%, demonstrating that even smaller LLMs can perform well when specialized fine-tuning is applied[48]. Additionally, fine-tuned LLMs have successfully learned radiation oncology-specific information and generated physician letters in required styles, with clinical experts rating the benefit at 3.44 out of 4[49]. These examples demonstrate the advantages of using fine-tuned models in various specialized medical domains.

Multilingual capabilities are also an important aspect of generalization and robustness for LLMs. In countries like the US, where people come from diverse language backgrounds, linguistic assistance is also important in clinical settings to support both patients and healthcare practitioners. Most current LLMs are developed based on English, performance differences between English and other languages on the same medical QA task can be as large as 0.30 in AUC[50]. For less common languages like Hebrew, LLM struggles to achieve a comparable performance to English[51]. As language barriers can impact patient care, improving LLM's multilingual capabilities is essential for LLMs to deliver consistent and accurate medical support across diverse languages.

## Operational Vulnerability

When closely examining generated content by LLMs, one of the most concerning issues discussed in the literature is the occurrence of hallucinations, where LLMs generate content that is inaccurate, inconsistent, or completely fabricated[11,52]. For instance, Jin et al, showed that ChatGPT generated fake article titles and PMIDs in order to use them as evidence to support its answers[53]. Hallucinations in LLMs can be broadly classified into two categories: intrinsic and extrinsic[52]. Intrinsic hallucinations occur when the generated text directly contradicts the input data, such as producing inconsistent and inaccurate output when processing clinical notes[11]. Extrinsic hallucinations refer to content that cannot be verified or refuted by the input source, which can occur when LLMs generate fabricated response when consulted for medical information[53] or references for medical literature[54].

In medical applications, these hallucinations can have serious consequences, such as the misinterpretation of clinical trial results[11,53] or the misclassification of patient data[55,56]. For instance, one recent study proposed to use GPT-4 to extract "helmet status" from patient clinical notes[56]. Although the model performed well in many cases, it exhibited hallucinations when it encountered negations like "unhelmeted," resulting in classification errors. Furthermore, LLM can generate self-conflicting answers when responding to healthcare-related questions. Agarwal et al. showed that when asked "Which foods cause the most allergies?", GPT-3.5 initially identified "fresh fruits and vegetables with high acidity" but later recommended "sticking to a diet of fresh, natural fruits," creating conflicting guidance[57].

A primary challenge in mitigating hallucinations is the difficulty of verifying the accuracy of generated content, especially when the training data is incomplete or when access to key sources (e.g., clinical trials) is restricted by copyright or other limitations[52]. In such cases, LLMs may fill gaps with inferential content, leading to extrinsic hallucinations that undermine the reliability of the presented information. Ji et al. explore both intrinsic and extrinsic hallucinations in medical, where they highlight how LLMs tend to generate factually incorrect content when faced with incomplete or ambiguous data, often fabricating plausible sounding but unfaithful responses[58]. Similar concerns with hallucinations in medical applications was found with the Google's Bard models[59]. Fabricated references and citations in the model's output show how hallucinations can mislead researchers and healthcare professionals.

Furthermore, the use of LLMs as end-to-end systems for tasks like clinical evidence summarization introduces additional complexities[53]. These processes often involve multiple steps, such as searching, screening, and appraising evidence, where errors in any step can cascade into the final output.

Adding to these challenges is the lack of reliable assessment for the factual accuracy of LLM-generated outputs. Existing automatic evaluation tools often do not interact well with expert human evaluations, particularly in tasks like medical evidence synthesis, where even minor inaccuracies can have significant repercussions[52]. Developing more effective and scalable methods for evaluating and verifying the accuracy of LLM outputs, such as self-reflection mechanisms[58], remains a critical area of research.

## Fairness and Bias

Fairness and bias are a major challenge for LLMs, with these models often reflecting and amplifying existing societal biases, such as those related to race, gender, and age. For example, in medical report generation, LLMs like GPT-3.5 and GPT-4 have been found to produce biased patient histories and racially skewed diagnoses, associating certain diseases disproportionately with specific racial groups[60]. These biases may stem from the imbalanced or insufficient data used in training these models, leading to discrepancies in diagnostic outcomes and patient care. In biomedicine, such biases can exacerbate existing healthcare disparities, disproportionately affecting marginalized groups and contributing to unequal treatment. The inconsistency of LLM outputs also reveals the bias inherent to the model[61]. For example, while some responses correctly identified that race is a social construct with no genetic basis, other responses from the same model contradicted this, incorrectly suggesting that race reflects subtle genetic influences.

Recognizing these issues, several studies emphasize the importance of using diverse evaluation methodologies and involving multiple stakeholders, such as physicians, health equity experts, and consumers, to identify biases that might otherwise remain undetected[62–64]. Pfohl et al. proposed EquityMedQA, a collection of adversarial datasets specifically designed to expose biases in LLM-generated responses to medical questions[62]. Alongside this, they introduced a multifactorial framework for evaluating LLMs based on six dimensions of bias, including inaccuracy for certain demographic groups, stereotypical language, and the omission of structural factors driving health inequities.

Addressing such biases in LLMs, particularly in biomedicine, is complex because bias can emerge at multiple stages—from data collection to model deployment. This issue is particularly relevant in healthcare, where biased AI-driven decisions can worsen health outcomes for vulnerable groups. Techniques such as adversarial learning, data augmentation, and representation learning have been proposed to mitigate these biases, but they often come with trade-offs in model performance. For instance, while some fairness interventions reduce bias in predictions, they can simultaneously lower the overall accuracy of the model or inadvertently introduce new biases[65,66]. Additionally, certain fairness metrics, such as equalized odds or demographic parity, may not fully capture the nuances of clinical decision-making, where the ranking of risk scores is critical for resource allocation and diagnosis[67]. Achieving a balance between fairness and performance remains an ongoing challenge, requiring continued research to ensure AI-driven medical tools are equitable and reliable across diverse patient populations.

## Ethical and Social Considerations

Recent research has revealed LLMs' concerning vulnerabilities to malicious manipulation that could potentially jeopardize patient safety and clinical decision-making practices. Both open source and proprietary LLMs are prone to manipulation in scenarios through targeted adversarial attacks[12,68]. These manipulative actions typically occur in two forms: crafted prompts (prompt-based attacks) and corrupted tuning data (fine-tuning attacks)[12]. By exploiting these techniques, malicious actors can manipulate LLM outputs to deliver targeted misinformation, and recommending incorrect or unnecessary medical procedures[12]. Using adversarial statements to deliberately change the weights, modified LLMs generate misinformation such as incorrect maximum dosage of drugs or other medical information, potentially leading to organ injury and drug misuse, with attack success rate reaching up to 99.7%[68]. Commonly used commercial models, including GPT-4 and GPT-3.5-turbo, are vulnerable to both prompt-based and fine-tuning attacks, leading to significant changes in suggesting unnecessary medical tests. For example, Yang et al., showed that the attacked models increased their CT scan suggestions from 48.76% to 90.05%, and MRI suggestions from 24.38% to 88.56%[12]. In one case, the model even suggested an MRI for an unconscious patient with a pacemaker, potentially causing serious harm. Hidden prompt injection attacks in medical imaging can manipulate the outputs of visual language models (VLM) like GPT-4o, with success rates as high as 70%, leading models to overlook critical conditions such as cancerous lesions during diagnosis[69].

The impact of these vulnerabilities in clinical settings is significant, extending far beyond typical concerns seen in general domains. The smooth flow of language in LLMs increases the likelihood of this danger since they can create explanations for incorrect conclusions that might even deceive healthcare experts[12,68]. Moreover, multiple works find that models that have been compromised by attacks might not exhibit decreases in their overall performance when tested against standard medical benchmarks, which makes identifying them particularly difficult[12,68]. As LLMs play a larger role in healthcare processes, from summarizing patient data to assisting with treatment choices, it is crucial to prioritize safeguard measures against malicious tampering. Prompt based attacks can be defended by making the system prompt of LLM applications transparent, reducing the likelihood of prompt injection by a third party. Fine-tuning attacks, however, currently lack reliable detection or robust defense methods. The best practice for healthcare practitioners is to use only LLMs from trusted sources to avoid potentially compromised models.

## Data Privacy and Security

As the development and implementation of medical LLMs often require private and domain-specific data, the use of LLMs in healthcare requires careful consideration of privacy and security needs. The misuse of healthcare LLMs, for example, could lead to unauthorized access to personally identifiable information (PII) or unintended consequences like membership inference, detecting a specific patient as a contributor to data[70]. LLMs are also commonly deployed for consumers, through accessible interfaces including web applications. Though these consumer-facing LLMs enhance accessibility and engagement, they introduce additional privacy and security concerns. LLMs that interact directly with users who may input sensitive information pose an increased risk of exposing PII, especially if data handling and storage practices are not sufficiently secure.

Regarding the deployment of LLMs in medical application, researchers have also emphasized the need for encryption, authentication, and access control mechanisms[71]. Despite some proposed solutions, the privacy and security risks associated with LLMs in healthcare have contributed to hesitancy and caution among many[72,73]. One of the concerns involves membership inference attacks (MIAs), where adversaries attempt to infer whether specific data was included in the model's training set[74]. Language Models trained on medical notes demonstrated that MIAs could reach an AUC of 0.90 when distinguishing whether patient belongs to the training set or not, exposing the vulnerability of models trained on healthcare data. Partially synthetic health data is also highly vulnerable to MIAs, with 82% of the patient data in one dataset being identified through the attack with high confidence.

Aside from deliberate attacks, the extensive capabilities of LLMs can result in inadvertent data memorization. For example, GPT-3.5 and GPT-4 models have demonstrated the ability to reproduce rows from a Diabetes dataset, which includes sensitive attributes like glucose levels, blood pressure, and BMI[75]. This memorization means that LLMs may unintentionally reveal private health information when prompted with unrelated or general questions, potentially breaching patient confidentiality. To effectively address these security and privacy challenges, further research must focus on enhancing both individual patient privacy and the overall security medical systems when applying LLMs.

## Legal Considerations

Medical LLMs should comply with a range of existing laws that govern data privacy, intellectual property, and medical device regulations. For instance, in the U.S., the Health Insurance Portability and Accountability Act (HIPAA) mandates strict standards for protecting patient data, meaning that LLMs handling sensitive health information must be designed with robust data security measures to prevent unauthorized access or data breaches[76]. Similarly, the General Data Protection Regulation (GDPR) in the European Union imposes requirements on data handling, including obtaining patient consent and ensuring data minimization, which can be particularly challenging for LLMs that rely on vast and diverse datasets for training[77].

As LLMs may generate outputs that mirror copyrighted medical literature or clinical guidelines, this also raises concerns about potential violations of intellectual property laws[78]. Regulatory agencies like the U.S. Food and Drug Administration (FDA) have yet to clearly define whether LLMs fall under the category of medical devices, which makes it uncertain what safety and efficacy standards apply to their deployment. Given these legal complexities, it is essential that medical LLMs are developed with these regulatory challenges in mind to ensure compliance.

The integration of LLMs into clinical practice could introduce liability considerations for physicians, due to the challenges of LLMs discussed previously in this article and beyond. These issues can lead to erroneous medical decisions, raising concerns about malpractice liability. The lack of legal precedents for LLMs in healthcare further creates uncertainty in how courts will handle cases involving LLM-influenced decisions[79]. Because current law typically holds physicians liable only if they deviate from the standard of care and cause an injury result, the legal environment naturally discourages the use of LLMs and restricts them to a supplementary role, limiting their potential to improve care[80]. It has been suggested that physicians should use LLMs to supplement, rather than replace, their clinical judgment to mitigate these risks, as courts may scrutinize their reliance on LLM outputs when evaluating negligence claims[81]. Challenges in verifying the reliability of LLM-generated recommendations also complicate adherence to the standard of care[79]. This presents a significant challenge to the widespread adoption and effective utilization of LLMs in healthcare.

# Conclusion

The growing integration of LLMs in healthcare holds significant potential for enhancing workflow, accuracy, and efficiency. However, acknowledging and addressing the challenges associated with LLMs is crucial for their safe and effective deployment in real-world medical settings. This article has outlined several key challenges, including operational vulnerabilities, ethical and social considerations, performance and assessment issues, and legal and regulatory compliances.

Addressing these challenges will be essential for ensuring the responsible and safe use of LLMs in medical applications. By developing strategies that mitigate risks, improve reliability, and establish clear guidelines, the medical community can build trust and accountability in the use of LLMs, ultimately enabling LLMs' full potential to benefit patient care.

# Acknowledgements

This work is supported by the NIH Intramural Research Program, National Library of Medicine.